# Cross-View Sequential Visual Localization with Spatio-Temporal Context Modeling for Autonomous Driving

Jiaping Wang, Shaobo Li, Zhen Wang*
School of Information Engineering, Chang'an University, P.R.China

***Abstract*—Continuous and reliable localization is essential for autonomous driving. Cross-view visual localization matches ground images with satellite maps, providing complementary localization cues for pipelines that depend on Global Navigation Satellite System (GNSS) signals and high-definition (HD) maps. Most existing cross-view visual localization methods process each frame independently, leaving temporal information underused and limiting accuracy under dynamic occlusion, illumination variation, and repetitive textures. This study proposes a temporal-context-enhanced framework for cross-view sequence visual localization. The proposed recurrent cross-frame module aggregates historical context from the previous state to enhance the coarse ground feature of each current frame. These enhanced features facilitate satellite candidate-region classification, while hierarchical fine-grained features enable precise local offset estimation. On the CVIS dataset, the proposed method reduces mean localization error from 3.80 m to 1.57 m and increases R@1 m from 8.14% to 40.22%. Direct transfer to KITTI-CVL achieves a mean error of 2.61 m, with target-domain fine-tuning further reducing the mean error to 2.27 m. Zero-shot field experiments on a real-world vehicle achieve a mean error of 2.84 m and R@5 m of 96.86%. These results demonstrate that temporal context enhancement significantly improves cross-view localization accuracy and supports robust deployment on public benchmarks and real-world roads.**

## I. INTRODUCTION

CONVENTIONAL autonomous driving (AD) systems typically follow a modular perception-prediction-planning pipeline, which requires centimeter-level localization to align vehicle states with high-definition (HD) maps [1]-[3]. Such localization relies on Global Navigation Satellite System/Inertial Navigation System (GNSS/INS) fusion, Light Detection and Ranging (LiDAR)-based mapping, and HD map matching, increasing cost and limiting scalability in large-scale or GNSS-degraded environments [4]. Although end-to-end AD has reduced the dependence on centimeter-level global poses, continuous, reliable, and lightweight localization remains crucial for global scene context and stable navigation [5]. As shown in Fig. 1, vision-based cross-view localization addresses this need by matching camera-captured ground images with satellite imagery, estimating the vehicle position on the satellite map without additional LiDAR sensing or HD map construction [6]-[9]. Because cameras are already standard on most autonomous vehicles, cross-view localization offers low sensor cost, flexible deployment, and stronger adaptability to environmental changes [10][11]. Compared with conventional LiDAR- and HD-map-dependent pipelines, this vision-only paradigm has emerged as a promising localization solution for outdoor AD in the presence of unstable GNSS signals [12]-[14].

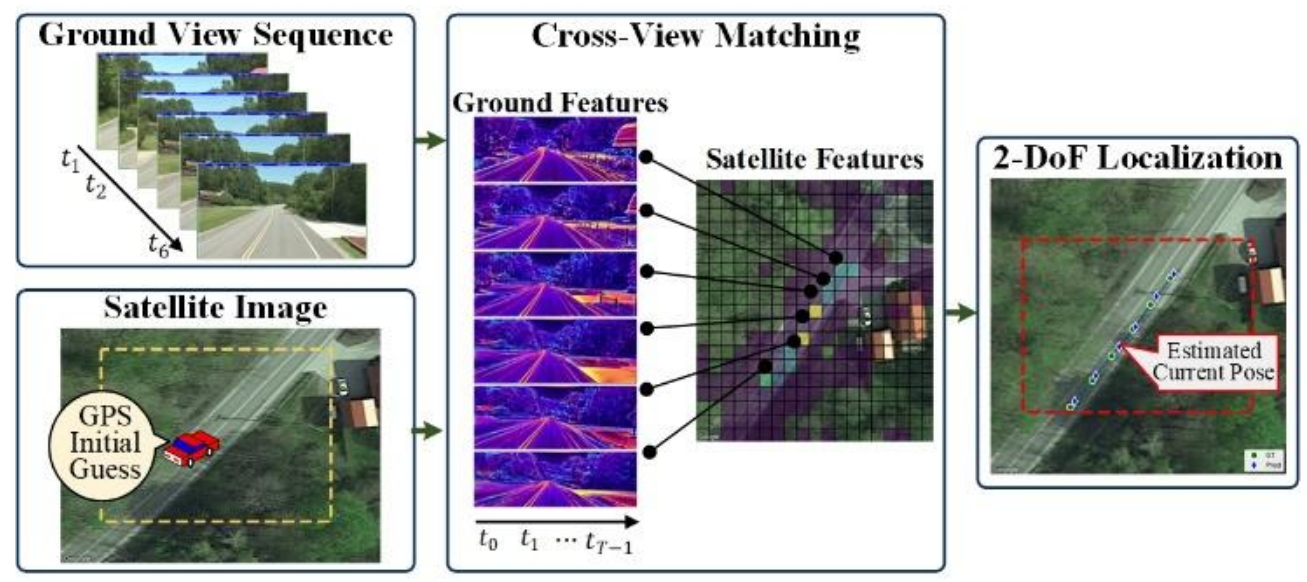

**Fig. 1.** Illustration of cross-view localization.

Most existing cross-view localization methods still treat road video as independent single-frame matching tasks [15]. Recent studies have improved single-frame localization through shared representation learning, cross-view fusion, local region alignment, and fine-grained position regression [16][17]. Shi et al. [18] reformulate retrieval-based localization as neural camera pose optimization, using geometric projection and a differentiable Levenberg–Marquardt module to iteratively estimate camera pose. Xia et al. [19] introduce dense uncertainty estimation, aligning ground and satellite features through learned uncertainty distributions for more robust metric localization. Deuser et al. [20] propose Sample4Geo, where two hard-negative sampling strategies improve contrastive cross-view retrieval on standard benchmarks. Wang et al. [21] design a correlation-aware homography estimator to recover precise geometric transformations and suppress spatial noise induced by viewpoint changes. Shore et al. [22] develop BEV-CV, projecting ground images into semantic Bird's-Eye View (BEV) representations to reduce cross-view geometry gaps under limited fields of view. Yuan et al. [23] build cross-view cross-attention between ground images and satellite maps, combining contextual spatial cues for local geometric matching beyond global descriptor retrieval. Xia et al. [24] further integrate geometric projection and alignment, establishing sparse or dense cross-view correspondences for direct vehicle-position estimation.

*Corresponding author: Zhen Wang (zhenwang@chd.edu.cn)

These single-frame methods improve cross-view localization accuracy across diverse scenarios, yet their independent matching paradigm discards temporal context from vehicle motion and continuous observations. In complex urban road scenes, vehicle-mounted cameras are vulnerable to dynamic occlusion, abrupt changes in illumination, and repetitive building textures [25]. Independent matching without historical context can produce ambiguous or incorrect frame-wise localization, limiting the reliability of single-frame localization in AD scenarios that require continuous position estimation [1].

Driven by the intrinsic limitations of single-frame matching, recent studies have shifted toward sequence-based cross-view localization using video frames. Sequence localization models contextual dependencies among historical observations, providing additional visual evidence for current-frame cross-view matching [26]. Shi et al. [27] present CVLNet, a framework that converts consecutive ground frames into BEV projections and matches satellite maps through inter-frame consistency in the projected view. Vyas et al. [28] introduce GAMa, which aggregates visual features from consecutive video clips and uses hierarchical search to improve sequence-to-map recall. Zhang et al. [29] propose a cross-view image sequence localization framework with temporal feature aggregation, capturing limited-field-of-view cues and mitigating feature loss caused by occlusion. Ghanem et al. [30] incorporate historical trajectories into ensemble learning to correct current-frame localization bias due to recent vehicle motion. Yuan et al. [31] develop a temporal-attention-based fine localization method, fusing spatiotemporal context from adjacent frames to reduce sequence localization errors. Deng et al. [32] present a self-supervised sequence framework that jointly encodes spatiotemporal features and samples frames through motion-aware dynamics, improving trajectory robustness in complex scenes.

Despite clear progress, existing sequence methods remain limited. Most approaches aggregate sequence information only at the feature level and do not inject historical context into the cross-view matching backbone, restricting the contribution of temporal cues to global matching and local fine localization. Although continuous frames can compensate for incomplete single-frame observations, current methods mainly emphasize holistic sequence aggregation and provide limited analysis of how temporal context drives fine-grained cross-view localization, thereby constraining deployment potential for continuous AD localization.

To address these limitations, this study proposes a temporal-context-enhanced framework for cross-view image sequence localization. The framework takes one satellite map and a continuous ground-image sequence as input, and introduces directional temporal propagation into the cross-view matching pipeline. Each current-frame feature adaptively aggregates historical scene cues from the previous recurrent state before candidate-region discrimination. The enhanced sequence features are then fed into candidate-region discrimination and local offset estimation modules, enabling fine-grained localization of consecutive frames on the satellite map.

The main contributions are as follows.

(1) This study proposes a cross-view sequence localization framework that integrates hierarchical feature representation, temporal context enhancement, and two-stage localization. The framework uses temporally enhanced coarse features for satellite candidate-region discrimination and fine-grained features containing local structure and texture cues for candidate-conditioned offset estimation, forming a continuous localization pipeline from coarse region localization to precise position prediction.

(2) This study designs a temporal context enhancement module for continuous time windows. At each timestamp, the current-frame feature retrieves historical information from the previous recurrent state through cross-frame attention, while a residual update preserves the current observation as the main representation. The recurrent structure progressively propagates historical context along the sequence, improving the discriminability of ground-view features.

(3) Extensive experiments on CVIS and KITTI-CVL demonstrate the effectiveness and generalization capability of the proposed method. On CVIS, the proposed method reduces mean distance error from 3.80 m to 1.57 m and improves R@1 m by 32.08 percentage points over SOTA. On KITTI-CVL, direct transfer reduces the mean error from 3.57 m to 2.61 m, while target-domain fine-tuning further reduces it to 2.27 m and improves R@1 m to 35.69%.

(4) A real-vehicle field experiment covers nine representative urban driving scenarios. The model uses low-accuracy GPS only to provide a coarse satellite-map prior and requires no additional training or fine-tuning. It achieves a mean error of 2.84 m, a median error of 2.92 m, and an R@5 m of 96.86%, while maintaining effective localization across the evaluated scenarios. In high-rise and occluded areas where RTK measurements drift, the proposed method still produces stable, road-aligned localization results.

The remainder is organized as follows. Section 2 details the proposed temporal matching localization framework and the temporal enhancement module. Section 3 reports the experimental design and result analysis, focusing on localization accuracy and robustness. Section 4 presents real-vehicle field experiments. Section 5 concludes the study.

## II. Methodology

### *A. Problem Statement*

This study addresses continuous cross-view matching localization for AD vehicles in complex urban environments, aiming to estimate vehicle trajectories on satellite maps from sequential ground observations. Given a ground-image sequence $S_g = \{I_{g,t}\}_{t=0}^{T-1}$, where $I_{g,t} \in \mathbb{R}^{H_g \times W_g \times 3}$ denotes the image captured by vehicle-mounted cameras at timestamp $t$, and a satellite map $I_s \in \mathbb{R}^{H_s \times W_s \times 3}$ covering the driving corridor without being centered on any queried vehicle position, the task estimates the planar vehicle positions $\{(x_t, y_t)\}_{t=0}^{T-1}$ in the satellite coordinate system. To solve this task, this section presents a temporal cross-view matching localization framework that models historical context among consecutive ground observations for continuous vehicle-position estimation. The following subsections first overview the overall architecture, then detail the local feature extraction module, temporal enhancement network, and two-stage localization module. For the reader's convenience, Table 1 summarizes the key notations used in this study.

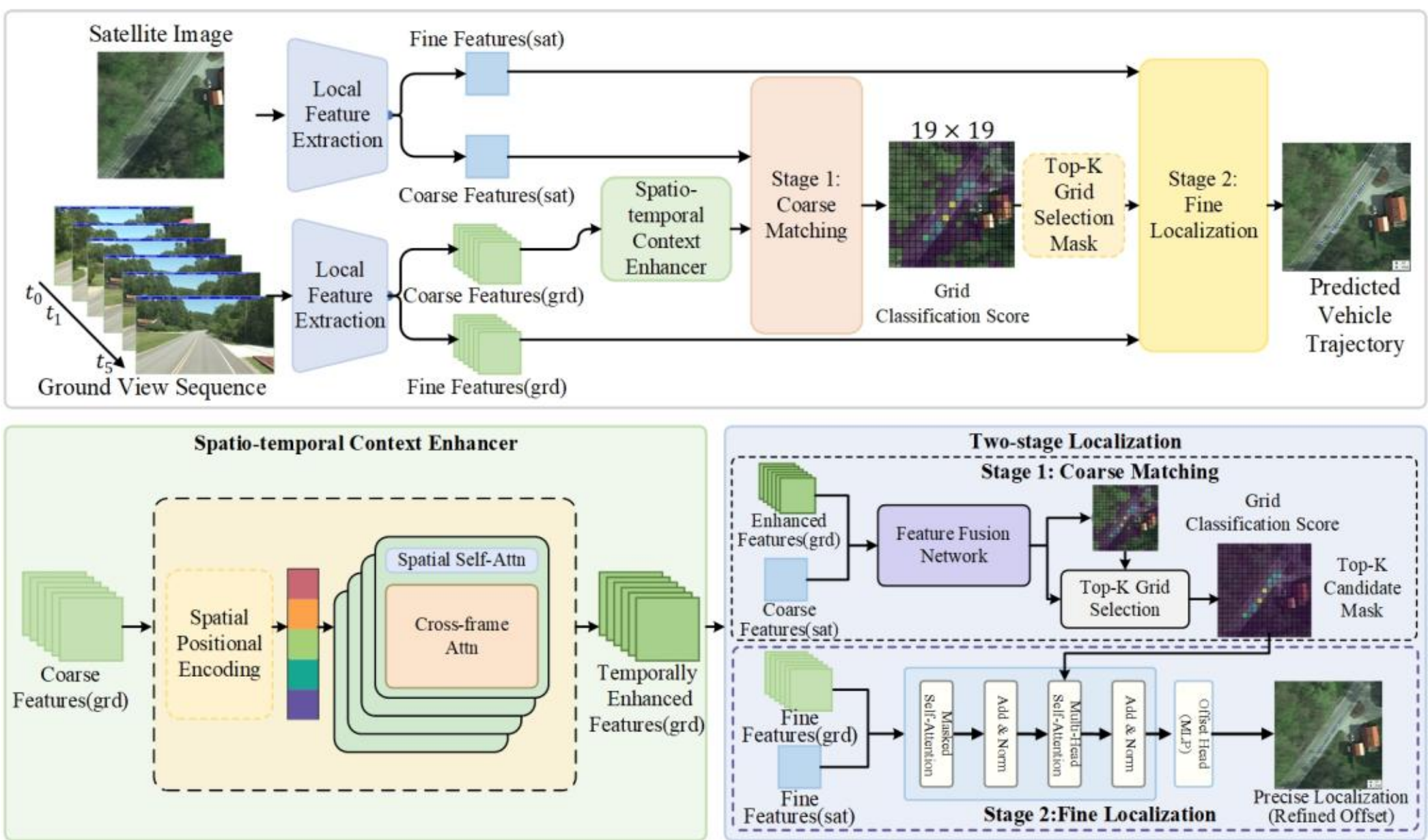

**Fig. 2.** Overview of the proposed sequential localization framework.

**Table 1.** Notation list.

| Symbol | Description |
|---|---|
| $I_s$, $I_{g,t}$ | Satellite image and ground-view image at time step $t$. |
| $F_s^c$, $F_s^f$ | Coarse-level and fine-level satellite features. |
| $F_{g,t}^c$, $F_{g,t}^f$ | Coarse-level and fine-level ground-view features of frame $t$. |
| $\tilde{F}_{g,t}^c$ | Temporally enhanced coarse-level ground-view feature of frame $t$. |
| $C_t$, $\mathcal{R}(\cdot)$ | Historical context aggregated from the previous recurrent state and the recursive update function. |
| $s_t$, $\mathcal{M}_t$ | Coarse satellite-grid classification scores and the corresponding top-K candidate mask. |
| $\Delta p_{t,n}$, $\hat{P}_t$ | Local offset predicted within grid $n$ and the final continuous predicted coordinate of frame $t$. |

*B. Sequential Localization Framework*

As shown in Fig. 2, the proposed localization framework consists of three main components. First, the local feature extraction module uses pretrained DINOv2 to extract coarse- and fine-grained representations from the satellite image and the ground-image sequence within a continuous time window. Second, the proposed temporal context enhancement module models historical context in coarse ground-sequence features, allowing each current frame to preserve its spatial layout while adaptively aggregating informative contextual cues from the historical state. Finally, the two-stage localization module performs feature fusion and mask-guided cascaded regression across the entire temporal clip, predicting continuous vehicle trajectory coordinates $\{(x_t, y_t)\}_{t=0}^{T-1}$. The following subsections detail each stage.

*C. Local Feature Extraction Module*

The local feature extraction module encodes one satellite map and six consecutive ground images to produce cross-view temporal representations for temporal enhancement and two-stage localization. As shown in Fig. 3, this study adopts pretrained DINOv2 with a Vision Transformer Base/14 (ViT-B/14) backbone to extract visual features from the satellite map $I_s$ and the temporally ordered ground sequence $S_g$, forming a continuous stream of ground-observation features.

Deep DINOv2 layers mainly encode abstract global semantics, whereas intermediate layers preserve stronger spatial layout, texture, and structural cues. This study therefore extracts hierarchical features from different backbone layers. The last backbone layer produces coarse-grained semantic features $F_s^c \in \mathbb{R}^{B \times C \times N \times N}$ and $F_{g,t}^c \in \mathbb{R}^{B \times C \times N \times N_1}$, where $F_s^c$ and $F_g^c = \{F_{g,t}^c\}_{t=0}^{T-1}$ denote the satellite and ground-sequence coarse features. To retain local texture and structural details, features from the 5th, 6th, 7th, and 8th intermediate layers are concatenated as fine-grained representations $F_s^f \in \mathbb{R}^{B \times C_1 \times N \times N}$ and $F_{g,t}^f \in \mathbb{R}^{B \times C_1 \times N \times N_1}$. Here, $B$ denotes the batch size, $C$ and $C_1$ denote channel dimensions, $T$ denotes the ground-sequence length, and $N$ and $N_1$ denote the feature-map sizes after patch partitioning. Unlike single-frame cross-view localization, this study preserves independent features for all six ground frames and explicitly organizes them along the temporal dimension, allowing each observation to retain local discriminability while supporting temporal dependency modeling in subsequent modules.

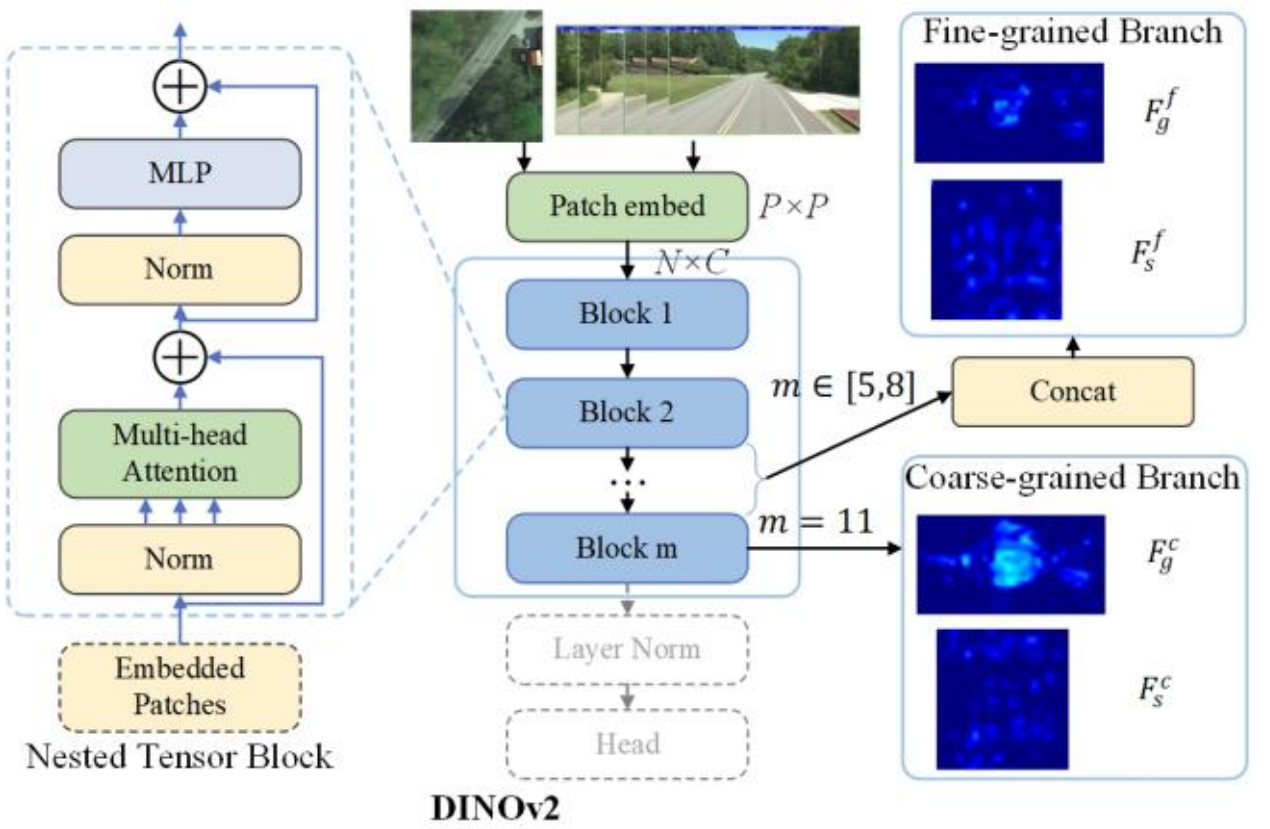

**Fig. 3.** Architecture of the DINOv2-based hierarchical feature extraction.

The satellite and ground hierarchical features are projected to a unified channel dimension $D$ through $1 \times 1$ convolution and then flattened along spatial dimensions. The projected ground-sequence features are further refined by the temporal enhancement module, and the enhanced sequence features are aggregated and matched in the two-stage localization module to estimate vehicle positions on the satellite map.

### *D. Spatio-temporal Context Enhancement*

Existing single-frame cross-view localization methods can achieve accurate matching in complex urban environments, yet they process ground observations independently at each timestamp and underuse temporal correlations between consecutive frames. In AD scenarios, adjacent ground images typically exhibit continuity in scene structure, viewpoint changes, and motion states, providing historical context that enables more stable current-frame cross-view matching.

As shown in Fig. 4, this study adopts a spatio-temporal cross-frame enhancement structure. The structure does not aggregate the entire sequence at once; instead, it updates ground-sequence features through state recurrence. Each frame is first enhanced by self-attention, and the first frame is then initialized as the historical state. For each subsequent frame, the current-frame feature serves as the target representation, and the previous recurrent state provides historical context. Historical information is progressively accumulated along the temporal direction, while the current frame remains the main component of the output state.

After spatial self-attention, the coarse ground feature of frame $t$ is denoted by $Z_{g,t}^c$. The recurrent state is initialized as $\tilde{F}_{g,0}^c = Z_{g,0}^c$. For frame $t \geq 1$, the recurrent state is updated as:

$$\tilde{F}_{g,t}^c = \mathcal{R}\left(Z_{g,t}^c, \tilde{F}_{g,t-1}^c\right), \quad (1)$$

where $\mathcal{R}(\cdot)$ denotes the recurrent cross-frame context update function. Unlike conventional sequence aggregation, $\mathcal{R}(\cdot)$ adopts an asymmetric Query-Key/Value design:

$$Q_t = Z_{g,t}^c + P^{cur}, \qquad K_t = \tilde{F}_{g,t-1}^c + P^{pre}, \qquad V_t = \tilde{F}_{g,t-1}^c \quad (2)$$

where $P^{cur}$ and $P^{pre}$ denote two-dimensional spatial positional encodings (SPE) for the current frame and the previous recurrent state, respectively. The asymmetric design gives a clear functional separation: the current frame $Z_{g,t}^c$ determines what information to retrieve, whereas the previous recurrent state $\tilde{F}^{c}_{g,t-1}$ provides retrievable historical context. The cross-frame attention (CTA) is formulated as:

$$C_t = Attn(Q_t, K_t, V_t) \tag{3}$$

where $C_t$ denotes historical context extracted from the previous recurrent state. The current frame is then fused with the historical context through a residual update:

$$\tilde{F}^{c}_{g,t} = Update\left(Z^{c}_{g,t}, C_t\right) \tag{4}$$

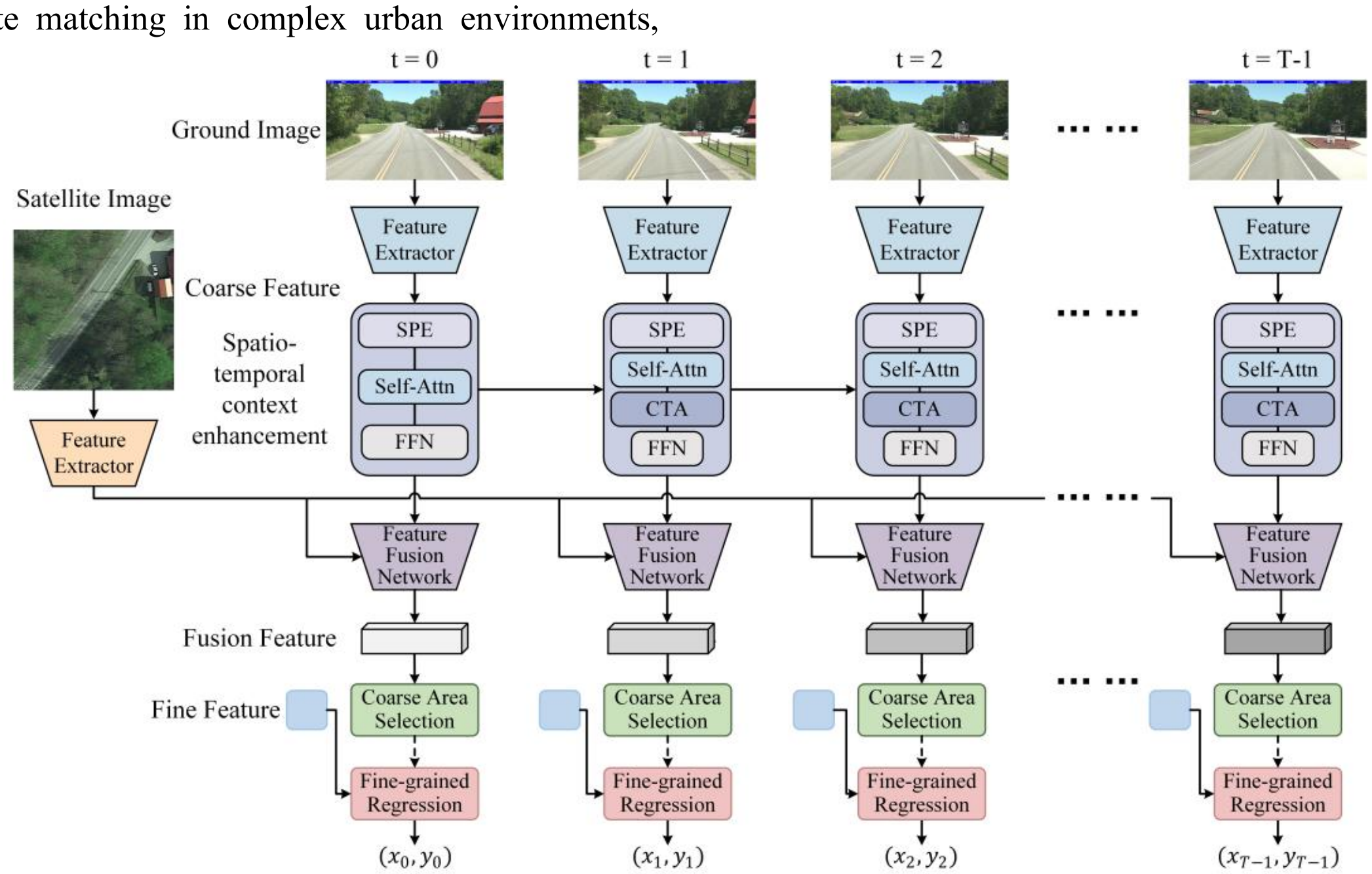

**Fig. 4.** Illustration of the spatio-temporal enhancement structure.

$Update(\cdot)$ consists of residual connections, normalization layers, and a feed-forward network. The updated state $\tilde{F}^{c}_{g,t}$ does not replace the current frame with historical information; instead, $\tilde{F}^{c}_{g,t}$ preserves $Z^{c}_{g,t}$ as the main representation and uses $C_t$ as complementary context. Therefore, the enhanced state maintains the spatial layout of frame $t$ and can be directly used for subsequent satellite-ground cross-view matching.

After recurrent historical-context aggregation, the context-enhanced ground sequence is obtained as:

$$\tilde{F}^{c}_{g} = \left\{\tilde{F}^{c}_{g,0}, \tilde{F}^{c}_{g,1}, \dots, \tilde{F}^{c}_{g,T-1}\right\} \tag{5}$$

The spatio-temporal context enhancement structure models historical information as a retrievable state rather than simply averaging or concatenating multi-frame observations. The current frame always serves as the Query and residual backbone, preventing accumulated history from suppressing discriminative cues in the current observation. The previous recurrent state serves as the Key and Value, enabling stable road structures and scene context to propagate frame by frame. In this way, historical cues from continuous observations are injected into coarse cross-view matching in a controlled manner, improving localization accuracy and prediction consistency in complex road environments.

*E. Two-stage Localization*

Although the spatio-temporal context enhancement module improves the discriminability of ground-sequence features, it does not alter the satellite search space. Instead, the enhanced features provide history-aware ground representations for subsequent cross-view matching. Therefore, this study adopts a two-stage localization strategy. Stage 1 uses coarse temporal features to select global candidate regions, identifying satellite grids likely to contain the vehicle. Stage 2 performs candidate-conditioned offset refinement, in which the candidate mask from Stage 1 guides fine-grained feature interactions and the corresponding local offsets refine the selected coarse grids.

**Stage 1.** As shown in Fig. 5, Stage 1 retrieves potential vehicle regions from the satellite map using coarse features from the satellite image and continuous ground observations. Unlike single-frame localization, the ground sequence first passes through spatio-temporal context enhancement and produces $\tilde{F}^{c}_{g,t}$, which allows historical observations to influence the satellite candidate distribution used for grid-level classification.

Specifically, let the recurrently enhanced ground coarse feature be $\tilde{F}^{c}_{g,t}$, and let the corresponding satellite coarse feature be $F^{c}_{s}$. Since the same satellite image is matched with each ground frame in the sequence, this study replicates the satellite coarse feature along the temporal dimension and fuses it with the enhanced ground feature at each timestamp. A location classification head then predicts the candidate-region distribution of frame $t$ over satellite grids:

$$s_t = \mathcal{H}_{cls}\left(\mathcal{F}_{cv}\left(F^{c}_{s}, \tilde{F}^{c}_{g,t}\right)\right). \tag{6}$$

where $\mathcal{F}_{cv}$ denotes the satellite-ground cross-view feature fusion module, $\mathcal{H}_{cls}$ denotes the candidate-region classification head, and $s_t \in \mathbb{R}^{N^2}$ denotes the classification scores of frame $t$ over $N^2$ satellite grids. The coarse candidate distribution provides global spatial constraints for the vehicle location and restricts subsequent fine-grained offset estimation to more reliable candidate regions.

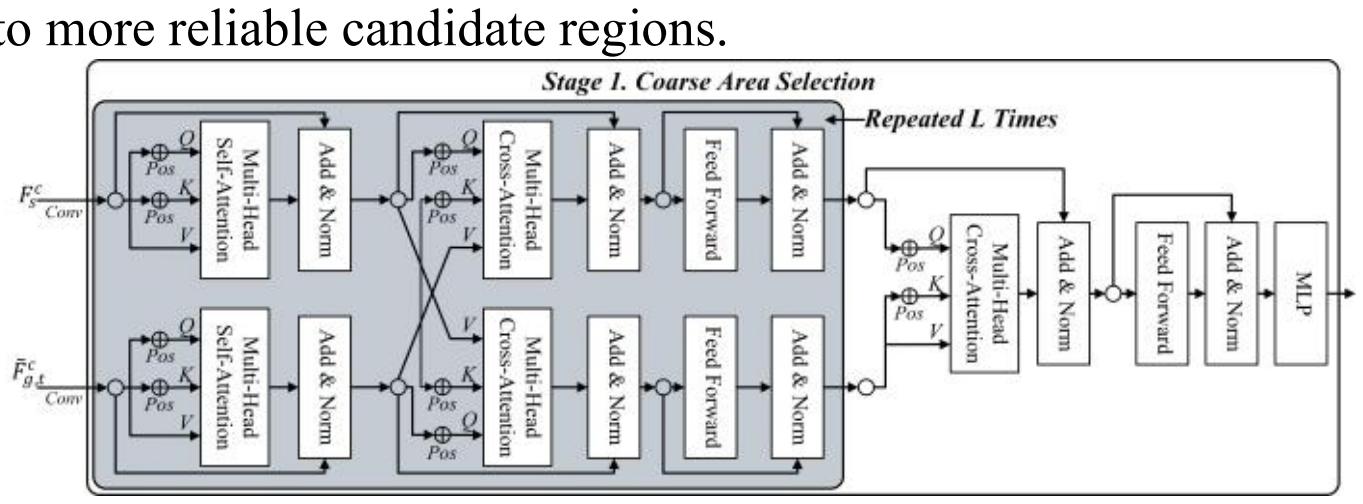

**Fig. 5.** Architecture of the dual-stream coarse localization network.

**Stage 2.** As shown in Fig. 6, after obtaining the coarse candidate-region distribution, this study constructs a candidate mask to guide fine-grained offset regression. The candidate mask is not the final location selection strategy; rather, it serves as a fine-grained feature constraint in Stage 2. Using only the Stage 1 top-1 grid for offset regression would prematurely fix the local uncertainty from coarse localization. Once the top-1 grid is biased by occlusion, repetitive textures, or adjacent road structures, the fine regressor cannot exploit complementary cues from other high-confidence candidates. In contrast, the top-K candidate mask preserves a set of high-confidence Stage 1 regions, avoiding large-scale background noise from the full satellite map while retaining local candidate context for fine offset estimation.

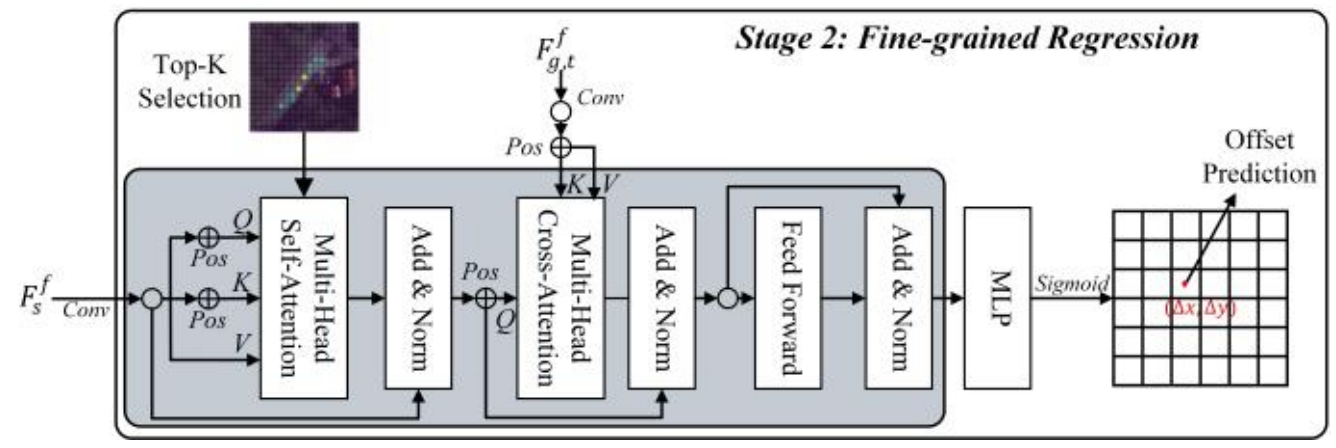

**Fig. 6.** Illustration of the mask-guided fine-grained localizer.

Specifically, according to the coarse classification scores, the top-K grids are selected to construct a binary candidate mask:

$$\mathcal{M}_{t,n} = \begin{cases} 1, & n \in TopK(s_t), \\ 0, & otherwise. \end{cases} \tag{7}$$

The candidate mask does not change the final output form of the model, but constrains the valid candidate regions during fine-grained offset decoding. The satellite and ground fine-grained features $(F^{f}_{s}, F^{f}_{g,t})$ extracted from intermediate layers are then used for local offset estimation:

$$\Delta p_{t,n} = \mathcal{H}_{reg}\left(F^{f}_{s}, F^{f}_{g,t}, \mathcal{M}_t\right), n \in \mathcal{M}_t \tag{8}$$

where $\Delta p_{t,n} = (\Delta x_{t,n}, \Delta y_{t,n})$ denotes the normalized local offset of frame $t$ relative to the $n$-th candidate grid, and $\mathcal{H}_{reg}(\cdot)$ denotes the candidate-guided offset regression module. Fine-grained offset prediction exploits road boundaries, local textures, and geometric details from intermediate structural features, compensating for the quantization error introduced by coarse grid classification. During inference, the grid with the highest coarse classification score is first selected as the final candidate region:

$$n_t^* = \underset{n}{\operatorname{argmax}}\, s_{t,n} \tag{9}$$

The continuous coordinate of frame $t$ on the satellite image is then decoded by combining the discrete grid coordinate with the corresponding local offset:

$$\widehat{P}_t = g(n_t^*) + \Delta p_{t,n_t^*} \tag{10}$$

where $g(n_t^*)$ denotes the two-dimensional coordinate of the $n_t^*$ -th satellite grid. The predicted coordinate is further converted into physical distance according to the satellite image resolution and ground sampling distance for localization-error evaluation.

As a local refiner, Stage 2 is not designed to recover arbitrary coarse-grid classification errors. Therefore, improving the Stage-1 candidate distribution is essential, and the temporal context module is introduced before Stage-1 classification to enhance candidate-region discrimination.

### *F. Training Objective*

The two localization stages are jointly trained using coarse-grid classification supervision and candidate-conditioned offset regression supervision. Considering the temporal nature of AD localization, the total objective aggregates classification and regression losses over all timestamps:

$$\mathcal{L}_{total} = \lambda_{cls}\mathcal{L}_{cls} + \lambda_{reg}\mathcal{L}_{reg} \tag{11}$$

where $\lambda_{cls}$ and $\lambda_{reg}$ balance the two learning objectives.

**Coarse Classification Loss.** In the coarse localization stage, the model predicts the satellite grid index containing the ground-truth vehicle position at each timestamp using temporally enhanced features. Since grid selection is a discrete classification task, cross-entropy loss supervises the sequence-level prediction and maximizes the probability of the ground-truth grid:

$$\mathcal{L}_{cls} = -\frac{1}{B \cdot T}\sum_{b=1}^{B}\sum_{t=0}^{T-1}\sum_{n=1}^{N^2} y_n^{(b,t)} \log\left(P_n^{b,t}\right) \tag{12}$$

where $N^2$ denotes the number of satellite grids. $y_n^{(b,t)}$ is the one-hot ground-truth label for sample $b$ at timestamp $t$, and $P_n^{b,t}$ denotes the predicted probability of grid $n$.

**Masked Regression Loss.** In the fine localization stage, the regression target is always defined on the ground-truth grid. The candidate mask determines whether the current candidate context covers the ground-truth grid. If the ground-truth grid is included in the candidate mask, the predicted offset at that grid is used to compute the regression loss. If the ground-truth grid is not covered, the regression loss of that sample is ignored to avoid supervising the local refiner under an invalid candidate context. The regression target is always indexed by the ground-truth grid, not by the Stage 1 top-1 prediction.

$$\mathcal{L}_{reg} = \frac{1}{B_{valid}}\sum_{b=1}^{B}\sum_{t=0}^{T-1} \mathbb{I}\left(M_{n_{gt}}^{b,t}\right) \cdot \left\| \hat{t}_{n_{gt}}^{b,t} - t_{gt}^{b,t} \right\|_2^2 \tag{13}$$

where $\mathbb{I}(\cdot)$ denotes the indicator function, and $B_{valid}$ denotes the number of valid regression samples in the current batch. $\hat{t}_{n_{gt}}^{(b)}$ and $t_{gt}^{(b)}$ denote the predicted normalized offset and ground-truth normalized offset, respectively. If the ground-truth grid is not selected by the candidate mask, the regression gradient at that timestamp is stopped.

## III. EXPERIMENTS AND RESULTS ANALYSIS

To evaluate the effectiveness of the proposed localization method, this section reports experiments on the CVIS [29] and KITTI-CVL [27] datasets. Section 3.1 describes the datasets and data splits; Section 3.2 defines the evaluation metrics for localization accuracy; and Section 3.3 details the implementation settings and hyperparameter configurations. Section 3.4 compares the proposed method against mainstream state-of-the-art (SOTA) methods across different datasets, and Section 3.5 presents ablation studies.

### *A. Datasets*

**CVIS.** The CVIS dataset contains real road-view sequences from Vermont, USA, and their corresponding satellite aerial images. It includes 118,549 street-view images captured by a front-facing vehicle camera with a 120° field of view. The dataset covers approximately 70% suburban roads and 30% urban areas, with about 30% of the data recording bidirectional trajectories along the same streets. To support sequential localization, adjacent street-view frames are spaced by about 8 m, and each sequence covers no more than 50 m. This constraint ensures that the entire trajectory segment lies within a single satellite map. Following this rule, CVIS provides 38,863 street-view sequence samples, each containing an average of 7 consecutive ground images. Each corresponding satellite map covers approximately 72.96 m×72.96 m with a ground resolution of about 0.114 m/pixel. To simulate localization uncertainty in real scenarios, the satellite map is not strictly centered on the street-view sequence and contains a random positional offset up to 5 m. This study splits CVIS into 24,872 training pairs, 6,218 validation pairs, and 7,773 test pairs. For a fair, fixed-length sequence input, this study samples six temporally ordered frames from each valid CVIS sequence during training and evaluation.

**KITTI-CVL.** KITTI-CVL extends the KITTI AD benchmark for cross-view localization. Shi et al. [27] use GPS coordinates from KITTI to retrieve high-resolution satellite imagery corresponding to street-view images. Each satellite map covers approximately 256 m × 256 m with a ground resolution of about 0.2 m/pixel. To adapt KITTI-CVL to sequential localization, this study serializes the original KITTI data using a strategy consistent with CVIS. Street-view images are sampled every 8 m to form a sequence $S = \{s_0, s_1, \dots, s_x\}$. Starting from $s_0$ , the distance to subsequent frames is accumulated, and $\{s_0, \dots, s_x\}$ is treated as one street-view sequence once the distance between $s_0$ and $s_x$ exceeds 50 m. The next sequence starts from the midpoint of the current sequence. Generated sequences with fewer than six street-view images are discarded. This process yields 1,077 street-view sequences, each paired with the satellite map corresponding to its midpoint. This study splits KITTI-CVL into 862 training pairs, 107 validation pairs, and 108 test pairs.

### *B. Evaluation metrics*

To comprehensively evaluate the proposed sequence localization model, this study follows common evaluation protocols for cross-view localization and measures performance from two aspects [23][31]: localization error statistics and distance-based recall. Model size and inference latency are also reported to assess computational cost and deployment efficiency, as summarized in Table 2.

**Table 2.** Evaluation metrics.

| Metric | Definition |
|---|---|
| Mean error | Average Euclidean distance error |
| Median error | Median Euclidean distance error |
| R@1 m | Percentage of predictions with error below 1 m |
| R@2 m | Percentage of predictions with error below 2 m |
| R@5 m | Percentage of predictions with error below 5 m |
| Params | Number of model parameters |
| Latency | Inference time for one input sequence |

### *C. Implementation details*

All experiments are implemented in PyTorch and trained on a single NVIDIA RTX 3090 GPU. For CVIS, ground images are resized from 1920 × 1080 pixels to 490 × 266 pixels, and satellite maps are resized from 640 × 640 pixels to 266 × 266 pixels during training and validation. For KITTI-CVL, ground images are resized from 1242 × 375 pixels to 490 × 154 pixels. Satellite maps are first center-cropped to 640 × 640 pixels and then resized to 266 × 266 pixels. Table 3 summarizes the network architecture and training hyperparameters.

**Table 3.** Network and training settings.

| Item | Setting |
|---|---|
| Feature dimension | C=768 |
| Temporal context module | 4 layers |
| Feature fusion module | 4 layers |
| Attention heads | 8 |
| Backbone learning rate | $10^{-6}$ |
| FFN hidden dimension | 1024 |
| Dropout | 0.1 |
| Satellite grid | 19×19 |
| Batch size | 4 |

### *D. Performance analysis*

This section quantitatively evaluates the proposed localization framework on the CVIS and KITTI-CVL datasets. To verify its effectiveness, this study compares the proposed method with three representative SOTA methods: CVML [19], CBSGV [23], and TACV [31]. Mean error, median error, and distance recall under different thresholds serve as the core metrics for sequence-based cross-view localization. The comparison results demonstrate that the proposed temporal framework consistently outperforms all three baselines across different scene clips.

**Evaluation on CVIS.** This subsection evaluates the proposed localization method on the CVIS dataset. Table 4 summarizes the overall performance of the proposed method and three baselines, while Fig. 7 reports distance-recall curves on the validation and test sets. Fig. 8 further visualizes localization results on different temporal clips, providing qualitative examples of trajectory-level localization behavior.

Table 4 reports the quantitative comparison on CVIS. The proposed method achieves the best localization performance on both validation and test sets. On the test set, the proposed method achieves a mean distance error of 1.57 m and a median distance error of 1.21 m, reducing the errors of the strongest baseline, TACV, by 2.23 m and 0.71 m, respectively. Compared with the variant without temporal enhancement, the full model reduces mean distance error from 5.92 m to 1.57 m, demonstrating the effectiveness of spatio-temporal context enhancement. Unlike TACV, which performs temporal aggregation on fused satellite-ground features, this study first enhances coarse ground features with historical context, then conducts cross-view feature interaction, and finally uses hierarchical features for candidate-region discrimination and local offset estimation. After replacing the TACV feature extractor with DINOv2, the test-set mean error decreases from 3.80 m to 2.84 m, and R@1 m improves from 8.14% to 15.95%, indicating that stronger visual representations improve sequential cross-view localization. Since TACV+DINOv2 already uses the same pretrained backbone, the remaining performance gap indicates that the improvement of the proposed method cannot be attributed only to backbone replacement. The placement of temporal enhancement, the use of hierarchical features, and the two-stage localization design also contribute to the final performance.

To reduce the influence of backbone differences, this study further replaces the original TACV feature extractor with DINOv2 while keeping its temporal aggregation strategy and localization structure unchanged. Although the full model introduces sequence modeling, its six-frame inference latency is 97.32 ms, lower than the 113.00 ms latency of TACV. This efficiency indicates that the proposed method achieves a favorable balance between localization accuracy and inference cost.

Fig. 7 shows the distance-recall curves of different methods on the CVIS validation and test sets. The proposed method consistently maintains higher recall across all distance thresholds, with the largest gains appearing under small-error thresholds. This improvement indicates that spatio-temporal context enhancement strengthens fine-grained position discrimination. As the threshold increases, the recall curve saturates after approximately 4 m and remains above all baselines, indicating reliable localization across different error tolerances.

**Table 4.** Localization performance comparison on the CVIS dataset. The best results are shown in bold, and the second-best results are underlined. "-" indicates that the corresponding metric is not reported in the original paper.

| Methods | Validation | | | | | Test | | | | | Params | Latency(ms) |
|---|---|---|---|---|---|---|---|---|---|---|---|---|
| | Mean | Median | R@1m | R@2m | R@5m | Mean | Median | R@1m | R@2m | R@5m | | |
| CVML[19] | 15.99 | 14.72 | - | - | - | 16.37 | 15.13 | - | - | - | 239.09 | 297.65 |
| CBSGV[23] | 10.97 | 7.52 | 6.95 | 20.86 | 37.16 | 11.39 | 7.89 | 6.76 | 19.57 | 34.12 | **27.09** | **21.86** |
| TACV[31] | 4.46 | 2.00 | 7.59 | 49.74 | 79.31 | 3.80 | 1.92 | 8.14 | 52.68 | 84.09 | 27.88 | 113.00 |
| TACV+DINOv2 | 4.35 | 1.61 | 14.84 | 62.54 | 83.18 | 2.84 | 1.48 | 15.95 | 70.19 | 91.77 | 183.83 | 103.53 |
| Ours w/o Temporal | 5.11 | 1.88 | 26.42 | 52.30 | 67.86 | 5.92 | 2.11 | 24.44 | 48.10 | 61.95 | 185.66 | 28.94 |
| Ours Full | **1.78** | **1.25** | **38.09** | **75.79** | **97.95** | **1.57** | **1.21** | **40.22** | **77.51** | **98.99** | 189.88 | 97.32 |

**Table 5.** Transfer and fine-tuning performance on the KITTI-CVL dataset. The best results are shown in bold, and the second-best results are underlined.

| Methods | Validation | | | | | Test | | | | |
|---|---|---|---|---|---|---|---|---|---|---|
| | Mean | Median | R@1m | R@2m | R@5m | Mean | Median | R@1m | R@2m | R@5m |
| CBSGV | 13.07 | 12.53 | 7.23 | 7.39 | 18.55 | 14.56 | 13.54 | 8.81 | 8.96 | 17.30 |
| TACV | 2.67 | 1.82 | 0.94 | 64.15 | 88.68 | 3.57 | 2.75 | 0.00 | 16.04 | 88.57 |
| Ours w/o Temporal | 12.52 | 9.43 | 11.32 | 16.82 | 18.71 | 14.38 | 13.85 | 9.59 | 13.68 | 16.67 |
| Ours Full | 1.42 | 1.12 | 43.55 | 78.46 | **98.27** | 2.61 | 1.97 | 26.42 | 51.10 | 89.47 |
| Ours + Fine-tuning | **1.14** | **0.85** | **58.33** | **86.48** | **98.27** | **2.27** | **1.55** | **35.69** | **63.36** | **92.92** |

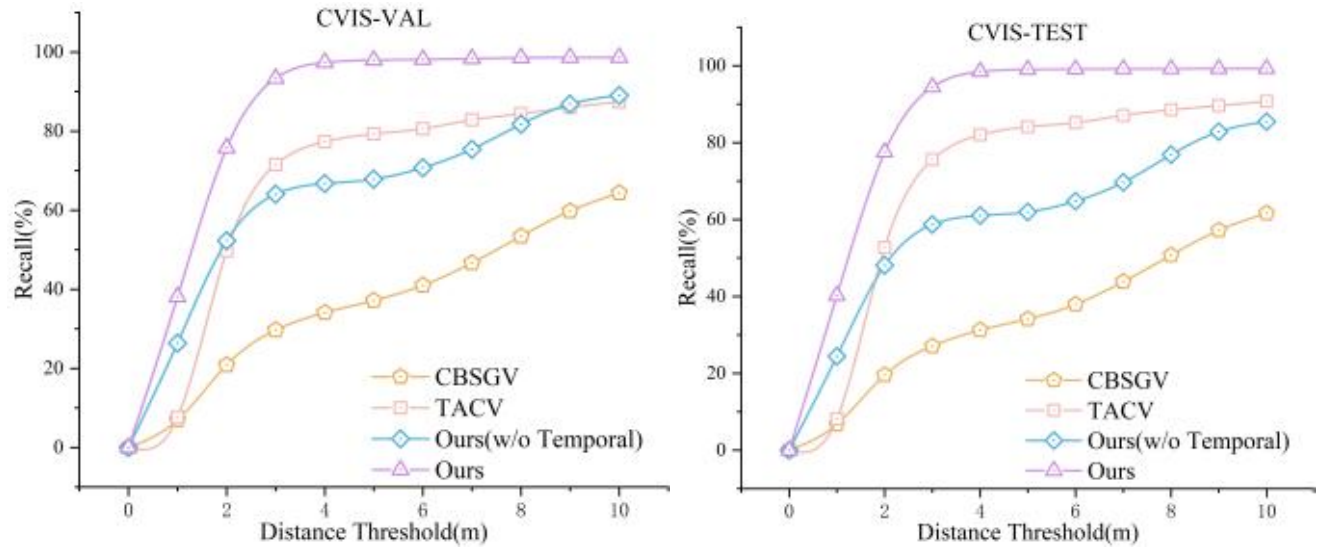

**Fig. 7.** Recall rate curves of different methods based on the CVIS dataset.

Fig. 8 visualizes localization results on two consecutive temporal clips. The predicted positions closely follow the ground-truth positions in spatial distribution. Together, Fig. 7 and Fig. 8 show that the proposed method improves distance recall and produces visually aligned localization results across consecutive frames.

**Fig. 8.** Visualization of sequential localization results for different temporal clips on the CVIS dataset. Green dots denote the GT trajectory, and blue diamonds denote our model's results.

**Evaluation on KITTI-CVL.** This subsection evaluates the transferability and target-domain adaptation capability of the proposed method on KITTI-CVL. Since the serialized KITTI-CVL dataset contains fewer samples than CVIS, this experiment serves as a supplementary transfer evaluation rather than a large-scale benchmark. Table 5 reports direct-transfer results and the performance obtained after fine-tuning the CVIS-pretrained model on KITTI-CVL. Fig. 9 shows distance-recall curves under varying thresholds, and Fig. 10 visualizes localization results on continuous temporal clips.

Table 5 shows that Ours Full consistently outperforms the baselines on KITTI-CVL. On the test set, Ours Full achieves a mean error of 2.61 m and a median error of 1.97 m, reducing the corresponding errors of TACV by 0.96 m and 0.78 m. Ours Full also achieves R@1 m and R@2 m of 26.42% and 51.10%, substantially outperforming TACV, whose R@1 m and R@2 m are 0.00% and 16.04%. Compared with Ours w/o Temporal, the full model reduces mean error from 14.38 m to 2.61 m, confirming the effectiveness of temporal context enhancement under cross-dataset evaluation. After fine-tuning on KITTI-CVL, Ours + Fine-tuning further reduces the test-set mean error to 2.27 m and improves R@1 m to 35.69%, indicating stronger adaptation to the target-domain distribution.

Fig. 9 presents the distance-recall curves on KITTI-CVL. The proposed method maintains a higher recall than all baselines across distance thresholds, with clear advantages at small error thresholds. Fine-tuning further improves recall at all thresholds, demonstrating that limited target-domain supervision helps the proposed framework adapt to the target-domain distribution and improve localization accuracy.

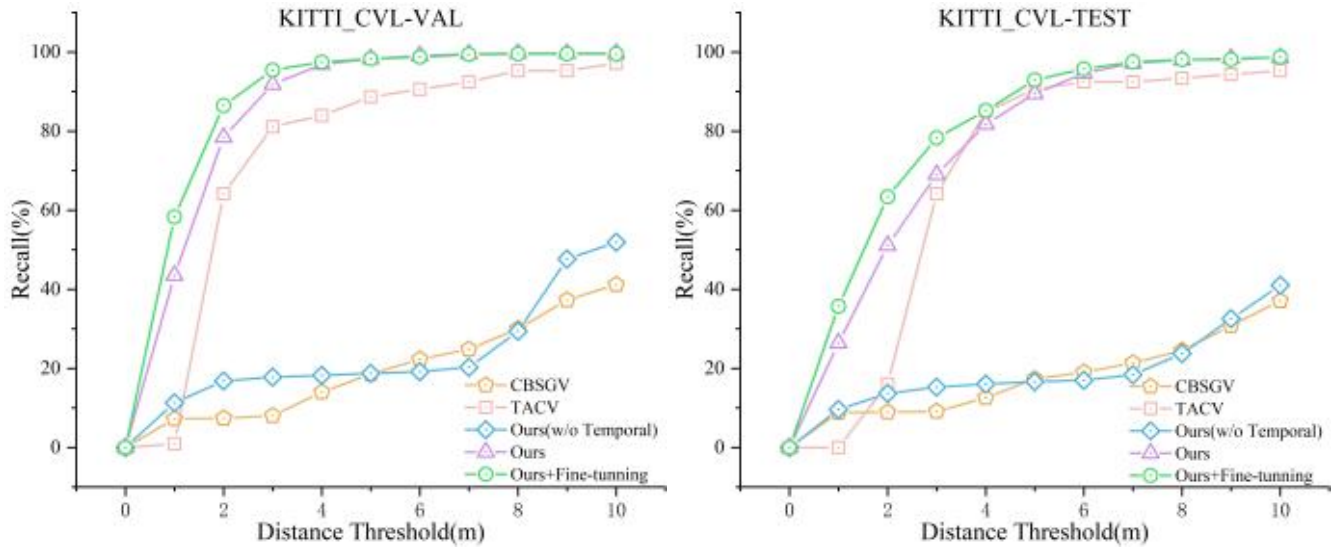

**Fig. 9.** Recall rate curves of different methods on the KITTI-CVL dataset.

Fig. 10 visualizes localization results on continuous KITTI-CVL temporal clips. The predicted positions are spatially close to the ground-truth positions in the visualized clips, indicating effective localization behavior in unseen road scenes. Together, Table 5, Fig. 9, and Fig. 10 demonstrate that the proposed method achieves strong cross-dataset performance in quantitative metrics and produces visually aligned trajectory predictions.

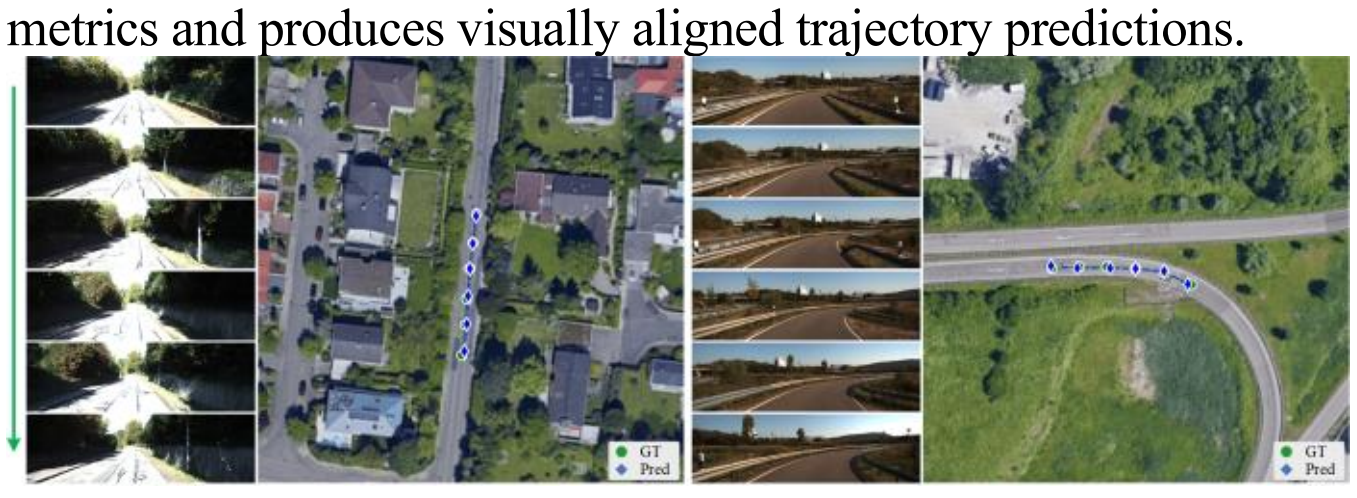

**Fig. 10.** Visualization of sequential localization results for different temporal clips on the KITTI-CVL dataset. Green dots denote GT, and blue diamonds denote our model's results.

### *E. Ablation study*

To analyze the contribution of each component, this study conducts stepwise ablation experiments on CVIS. As reported in Table 6, Baseline uses only DINOv2 features for cross-view matching; +Multi-level Features further introduces a multi-level feature representation; and +Temporal Context adds the proposed temporal context enhancement module.

Table 6 shows that using only final-layer DINOv2 features gives a test-set mean error of 12.25 m, indicating that a single semantic feature level cannot simultaneously support coarse candidate-region recognition and fine offset estimation. Introducing hierarchical features reduces the mean error from 12.25 m to 5.92 m and improves R@1 m from 6.37% to 24.44%. This improvement indicates that features at different depths provide complementary information for the two localization stages: deep features offer stronger semantic discrimination for coarse satellite-region classification, whereas intermediate features preserve richer local structures and textures for fine offset refinement. To further analyze the internal components of the recurrent cross-frame context enhancement module, this study constructs a position-aware variant without cross-frame aggregation. This variant removes cross-frame attention between the current frame and the previous recurrent state, injects only two-dimensional spatial positional encoding into the current coarse ground feature, and produces the output through the same residual feed-forward update as the full module. The position-aware spatial update reduces the mean error to 4.96 m, indicating that spatial structural priors improve the representation of ground features. Adding temporal context enhancement reduces the mean error to 1.57 m and increases R@1 m and R@5 m to 40.22% and 98.99%, respectively. These results demonstrate that historical observations strengthen the discriminability of coarse ground representations and generate more reliable satellite candidate distributions for subsequent refinement. Overall, hierarchical feature utilization improves compatibility between feature abstraction levels and stage-specific localization objectives, while temporal context reduces frame-level ambiguity by aggregating historical evidence. Their complementary effects jointly improve final localization performance.

To analyze how temporal context affects fine localization, this study further examines the coverage and probability distribution of Stage 1 candidate regions. Since Stage 2 performs offset regression only within the candidate regions selected by Stage 1, the quality of the coarse candidate distribution directly affects fine localization. Temporal context can improve final localization accuracy only when it helps the model retain the ground-truth region inside the candidate mask.

Table 7 reports ground-truth grid coverage under different top-K settings. Compared with the model without temporal context, the full model achieves higher candidate coverage for all K values. This improvement is especially important when only a small number of candidates are retained, since visually similar satellite regions can easily be confused with the true location. Under the top-64 setting, the full model uses only 17.73% of the 19×19 search space while covering 99.98% of ground-truth grids. These results indicate that temporal context improves the reliability of candidate-region discrimination before fine offset regression.

**Table 7.** Stage-1 candidate coverage on the CVIS dataset.

| top-K | Candidate Ratio | w/o Temporal | Full |
|---|---|---|---|
| top-1 | 0.28% | 43.55 | 69.26 |
| top-5 | 1.39% | 83.33 | 98.92 |
| top-10 | 2.77% | 93.64 | 99.76 |
| top-32 | 8.86% | 99.75 | 99.95 |
| top-64 | 17.73% | 99.95 | 99.98 |

**Table 6.** Ablation study on CVIS dataset.

| Model | Multi-level Feat. | Pos. Enc. | Temporal Attn. | Mean | Median | R@1m | R@2m | R@5m |
|---|---|---|---|---|---|---|---|---|
| Baseline | ✗ | ✗ | ✗ | 12.25 | 12.77 | 6.37 | 11.91 | 16.78 |
| + Multi-level Features | ✓ | ✗ | ✗ | 5.92 | 2.11 | 24.44 | 48.10 | 61.95 |
| + Position-aware Update | ✓ | ✓ | ✗ | 4.96 | 1.79 | 28.00 | 53.97 | 68.46 |
| + Temporal Context | ✓ | ✓ | ✓ | 1.57 | 1.21 | 40.22 | 77.51 | 98.99 |

Fig. 11 further visualizes the Stage 1 probability distributions. Without temporal context, the response distribution tends to spread across multiple visually similar road regions. After the temporal context is introduced, high-confidence responses become more concentrated around the reference region. These observations indicate that the proposed temporal module does not serve as post-processing or output smoothing; instead, it refines the satellite candidate distribution prior to Stage 2 and provides a more reliable spatial mask for local offset estimation.

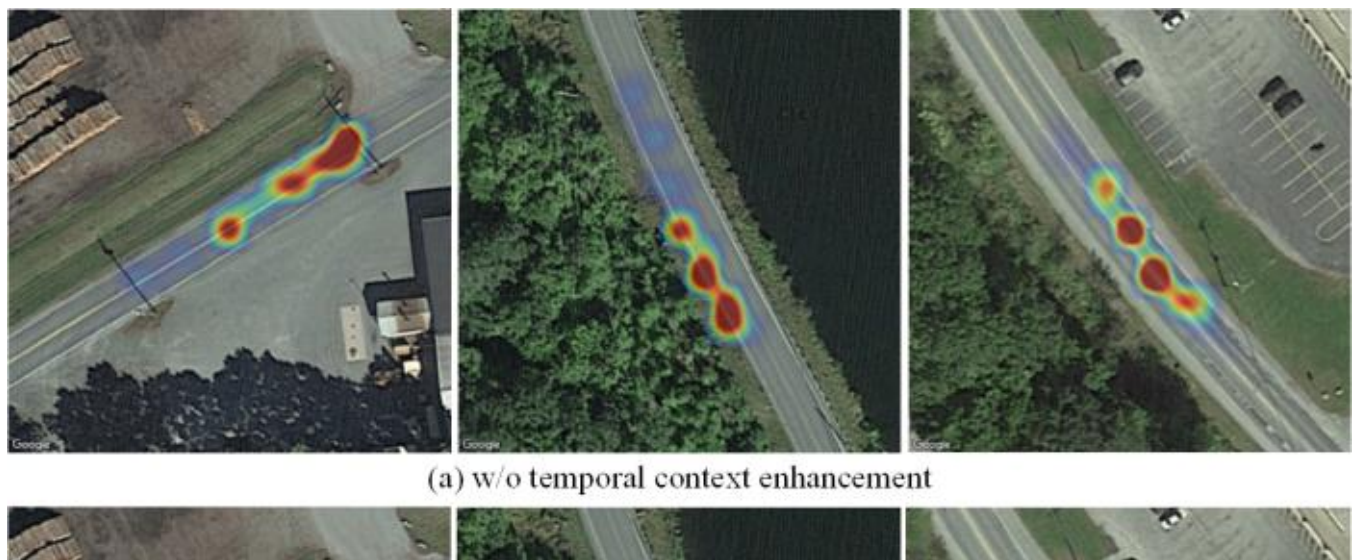

(a) w/o temporal context enhancement

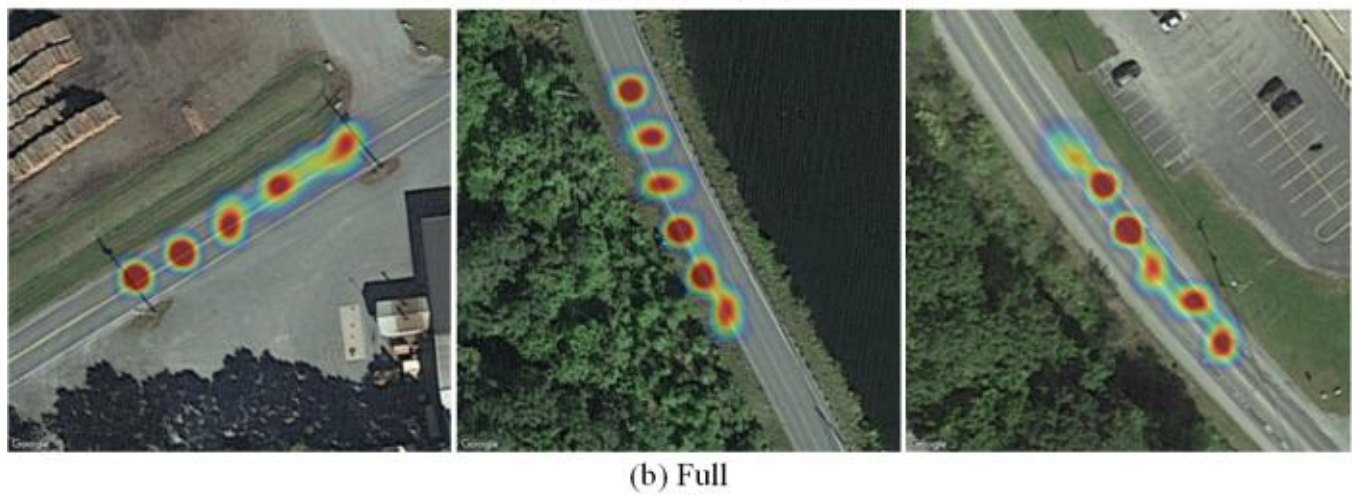

(b) Full

**Fig. 11.** Effect of temporal context on Stage-1 candidate distributions.

## IV. REAL-WORLD VEHICLE EXPERIMENT

To further verify applicability under real-vehicle conditions, this study conducts field experiments on a real vehicle. Unlike public-dataset experiments, the real-vehicle experiment uses low-accuracy GPS measurements to define satellite-map crop centers, while high-accuracy GPS serves as ground truth. The model directly uses parameters learned from public datasets, without training, fine-tuning, or updating on real-vehicle data, thereby evaluating zero-shot deployment capability in real-vehicle applications.

### *A. Experimental settings*

Fig. 12 shows the vehicle used in the experiments. The vehicle is equipped with a front-facing camera, a high-accuracy GPS device, and a low-accuracy GPS device to synchronously collect road images and vehicle positioning information. As reported in Table 8, the front-facing camera provides continuous road-view images during driving; the high-accuracy GPS device provides real-time kinematic (RTK)-based GT trajectories for localization-error evaluation; and the low-accuracy GPS device offers coarse positioning from ordinary onboard localization equipment and defines the satellite-map crop center.

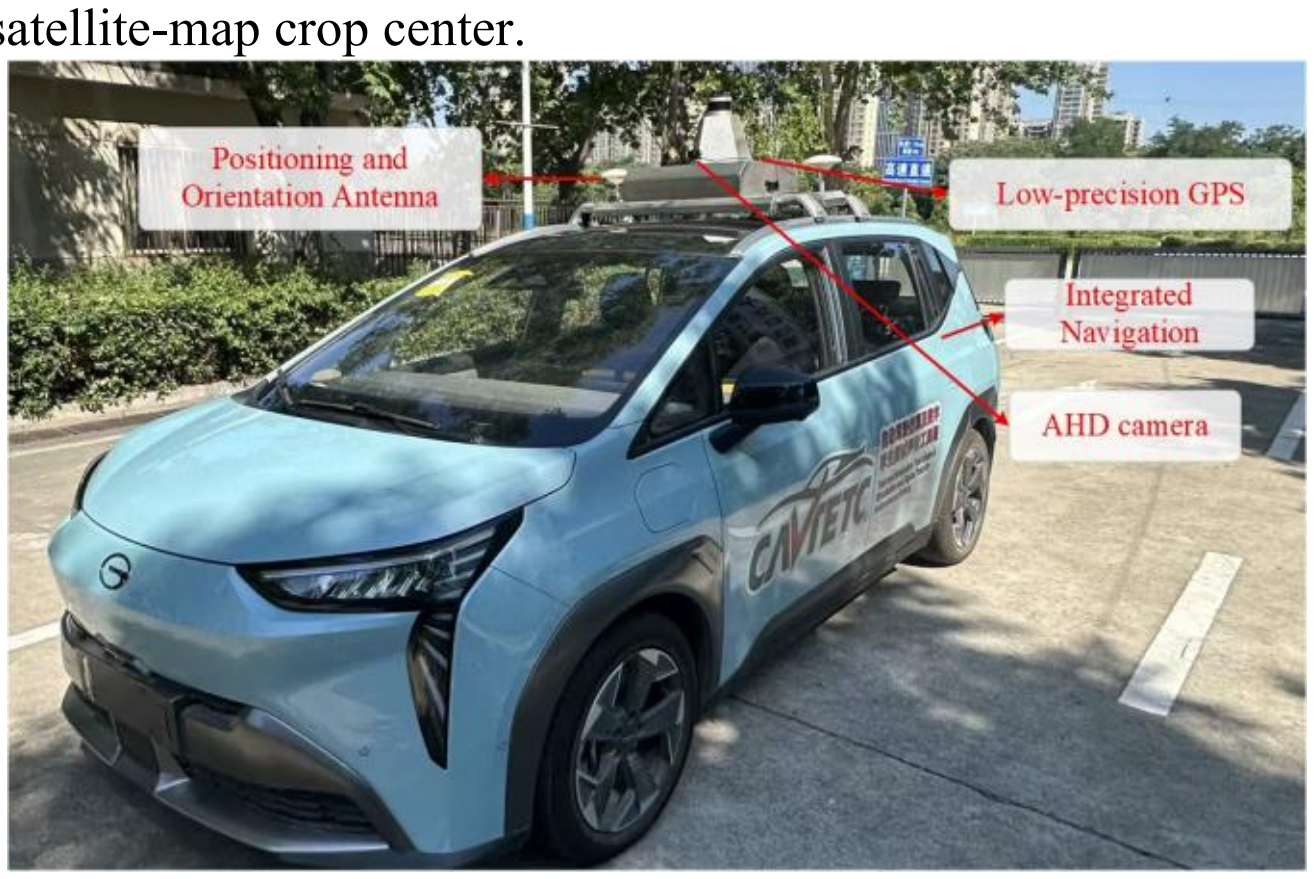

**Fig. 12.** Real-world vehicle platform and sensor setup.

As shown in Fig. 13, the driving route in the field experiments covers diverse real urban road scenarios, including arterial roads, intersections, dusk scenes, roundabouts, elevated roads, uphill and downhill segments, and side roads under viaducts. These scenes contain structural road changes, occlusions, elevation variations, and complex traffic conditions, providing a realistic testbed for cross-view visual localization in real-world driving.

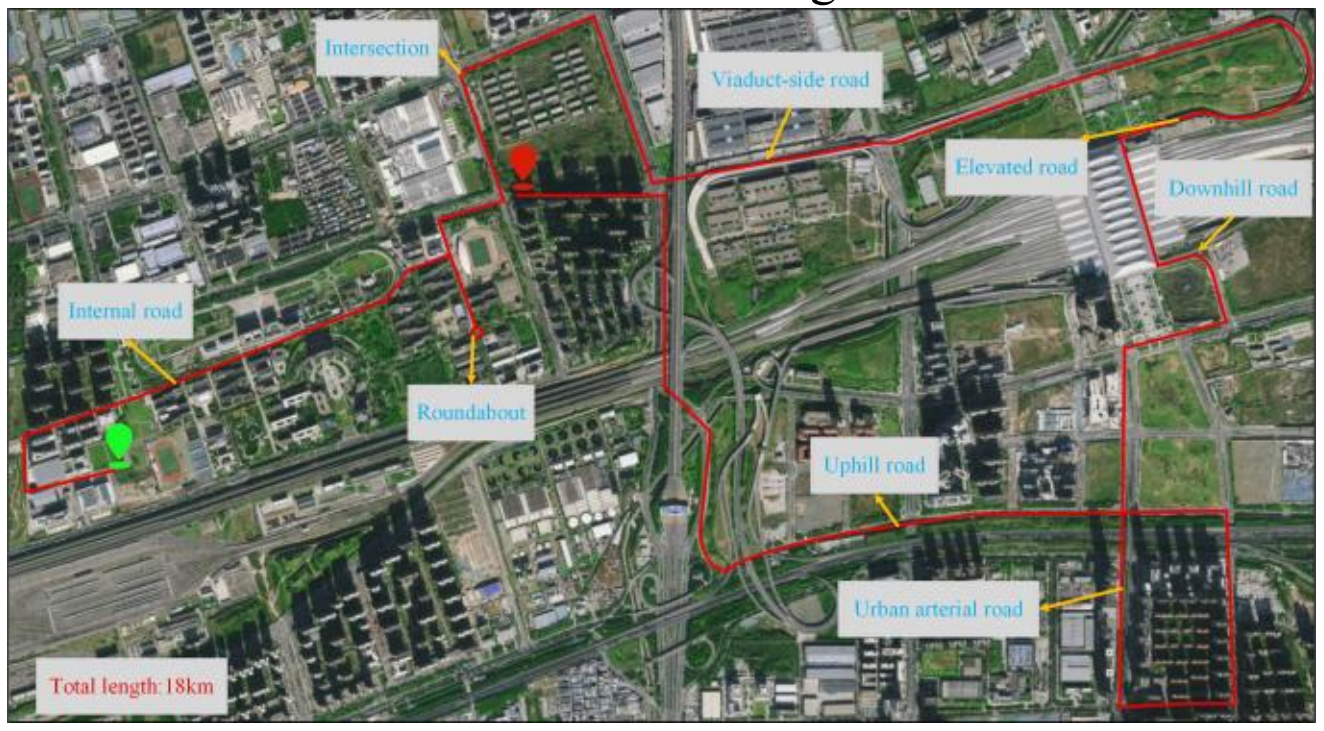

**Fig. 13.** Driving route in the field experiments.

For each ground-image sequence, this study uses the low-accuracy GPS position of the center frame as the crop center for the satellite image and retrieves the corresponding satellite image from Google Maps. Satellite images are acquired with zoom=20 and scale=2, and each crop is resized to 640 × 640 pixels. At this zoom level, each satellite image covers approximately 78.9 m × 78.9 m, encompassing the low-accuracy GPS error range and the surrounding road context.

**Table 8.** Detailed configuration of the real-world vehicle platform.

| Item | Model/Type | Main specification | Sampling rate |
|---|---|---|---|
| Front-view Camera | AHD | 1280×720 resolution | 25 FPS |
| Low-precision GPS | WTGAHRS3-485 | GNSS/IMU module;<br>meter-level positioning accuracy | 10 Hz |
| High-precision GPS | CHCNAV CGI430 | RTK-capable GNSS/INS;<br>horizontal RTK accuracy of 1 cm+1 ppm | 50 Hz |

Table 9. Scene-wise localization performance on real-vehicle data.

| Scene | Sequences | Mean | Median | R@1m | R@2m | R@5m |
|---|---|---|---|---|---|---|
| Urban arterial road | 148 | 1.78 | 1.63 | 16.67 | 65.88 | 99.89 |
| Intersection | 121 | 3.02 | 3.24 | 14.46 | 28.65 | 93.81 |
| Internal road | 125 | 1.38 | 1.12 | 43.20 | 83.73 | 100.00 |
| Roundabout | 15 | 2.30 | 2.10 | 6.66 | 47.78 | 96.67 |
| Dusk scene | 86 | 2.86 | 2.89 | 9.11 | 28.68 | 96.51 |
| Elevated road | 335 | 3.67 | 3.63 | 0.19 | 3.58 | 94.77 |
| Viaduct-side road | 159 | 2.95 | 2.89 | 1.78 | 22.32 | 98.95 |
| Downhill road | 33 | 3.33 | 3.32 | 3.53 | 8.59 | 97.47 |
| Uphill road | 9 | 4.21 | 4.37 | 1.85 | 5.55 | 87.04 |
| All scenes | 1031 | 2.84 | 2.92 | 10.65 | 30.98 | 96.86 |

*B. Scene-wise quantitative and qualitative results*

This subsection evaluates real-vehicle localization performance. The experiment directly uses the model trained on the CVIS dataset and applies no additional training, fine-tuning, or parameter update on real-vehicle data. The Euclidean distance between model outputs and GT is used to compute mean error, median error, and distance recall. To analyze performance across real road conditions, the experiments are divided into nine typical scenarios: urban arterial roads, intersections, internal roads, roundabouts, dusk scenes, elevated roads, side roads under viaducts, downhill segments, and uphill segments. Table 9 reports the scene-wise quantitative results.

Table 9 shows that the proposed method achieves effective localization performance in real-vehicle application. Across all scenarios, our model achieves a mean error of 2.84 m, a median error of 2.92 m, and R@5 m of 96.86%, demonstrating effective cross-view localization under real-road conditions. The best performance is achieved in the internal-road scenario, with a mean error of 1.38 m, a median error of 1.12 m, R@1 m of 43.20%, and R@5 m of 100.00%. In the urban arterial roads and side roads under viaducts scenarios, our model also yields strong results, with mean errors of 1.78 m and 2.95 m, respectively. These results indicate that clear road boundaries and persistent scene structures provide useful cross-view cues for satellite-ground matching.

In contrast, our model produces relatively greater localization errors in intersections and elevated roads. Our model achieves a mean error of 3.02 m and R@1 m of 14.46% in the intersection scenario, suggesting that complex road topology, multi-directional lanes, and dynamic traffic interference increase cross-view matching ambiguity. In the elevated-road scenario, our model achieves a mean error of 3.67 m and R@5 m of 94.77%, but R@1 m and R@2 m remain low. The main reason is that elevated roads contain similar road shapes and lane textures, with fewer distinguishable surrounding landmarks, causing candidate-region confusion and less accurate local matching. In the downhill and uphill scenarios, our model achieves mean errors of 3.33 m and 4.21 m, respectively, indicating that slope-induced viewpoint changes affect localization but still allow high recall within the 5 m threshold.

Fig. 14 visualizes localization results across different real-vehicle scenarios. For internal roads, urban arterial roads, and side roads under viaducts, our model's results are generally close to the GT, demonstrating the strong usability in the real world. In roundabouts, elevated roads, and sloped segments, local trajectory deviations appear, but our visual localization results still follow the overall road geometry. Notably, in the uphill scenario, our model still produces reasonable localization results despite dynamic occlusion from windshield wiper motion, suggesting the robustness to local visual interference during real driving.

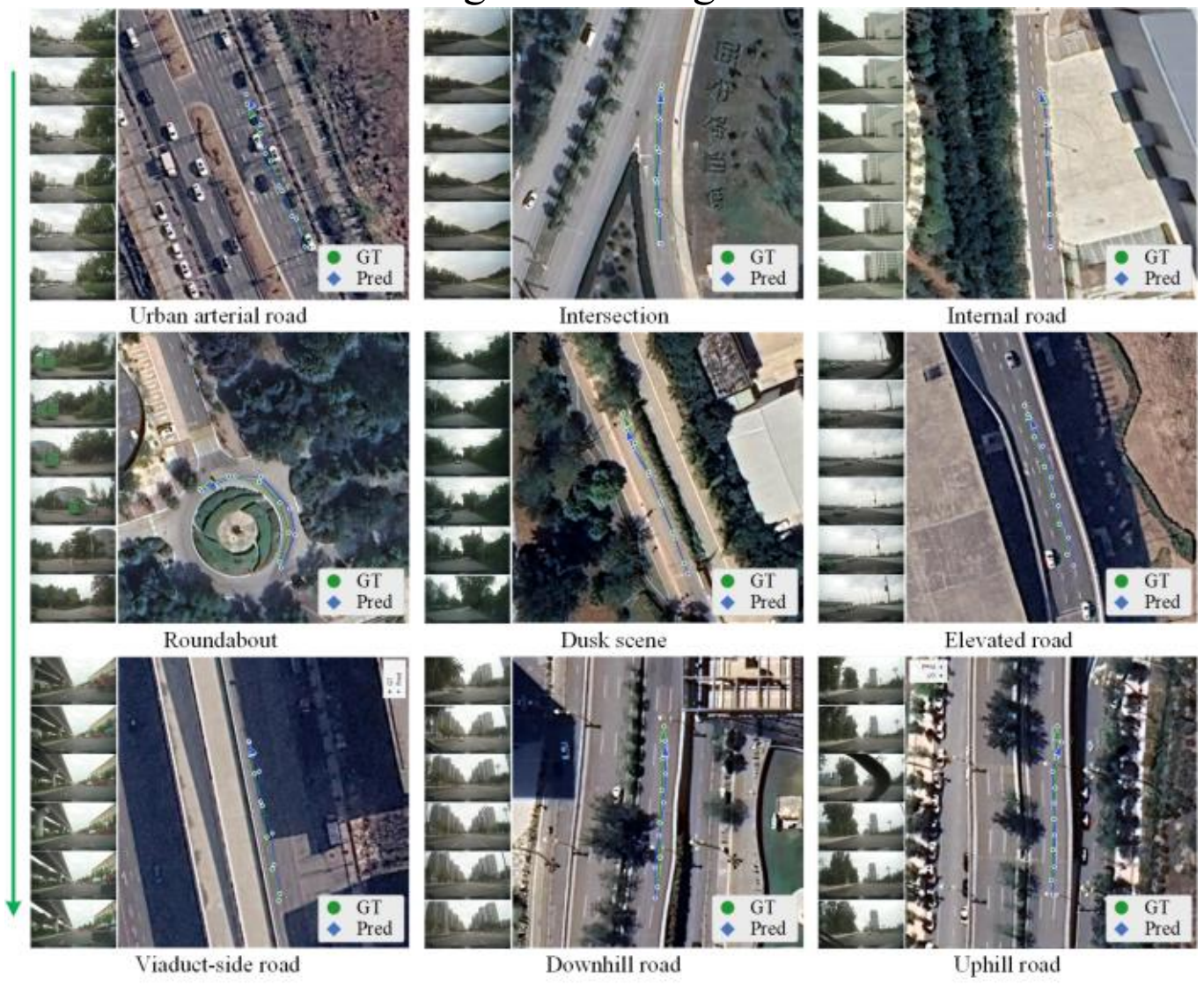

**Fig. 14.** Localization results in different real-world driving scenarios. Green dots denote GT, and blue diamonds denote our model's results.

Fig. 15 further shows two special cases. Under some severe scenarios (Severely obstructed by buildings and trees), RTK GPS is unable to provide correct localization results. Our visual localization model is still able to provide precise localization results.

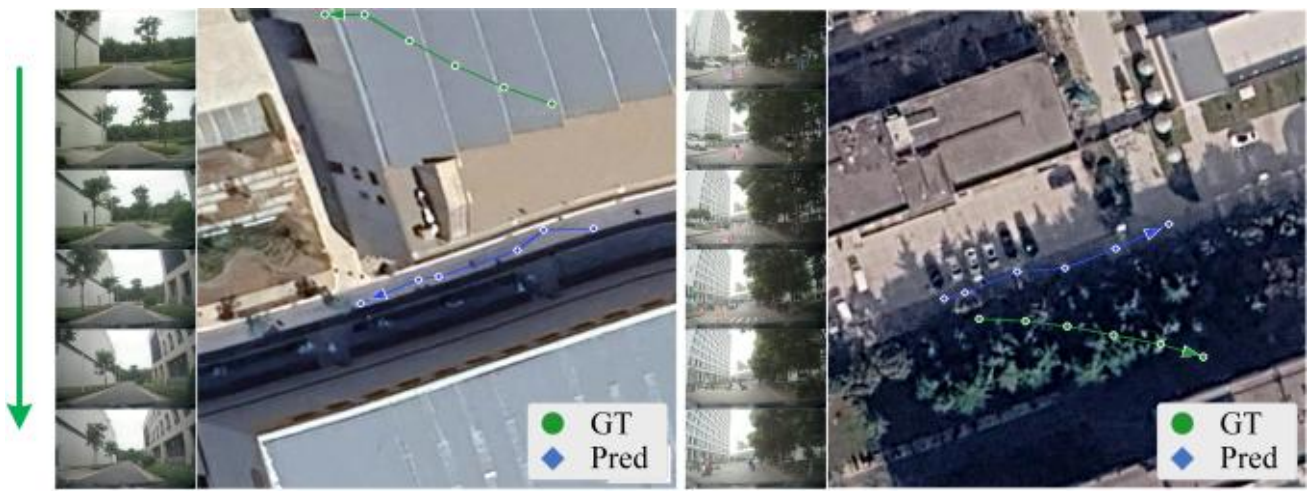

**Fig. 15.** Qualitative cases with local inconsistencies between RTK-based GT and visible road geometry. Green dots denote results from RTK, and blue diamonds denote our model's results.

Overall, Table 9, Fig. 14, and Fig. 15 show that the proposed method performs continuous cross-view localization on real roads using only low-accuracy GPS and without training on real-vehicle data, suggesting practical deployment potential.

## V. Conclusion

This study proposes a temporal cross-view localization method for AD. The method recurrently enhances ground-view features, using the current frame as the Query and the previous state as the Key and Value, allowing each frame to retrieve historical scene cues while preserving its spatial layout. The enhanced features support satellite-grid candidate discrimination and local offset estimation, improving continuous-frame localization accuracy and robustness. Extensive experiments demonstrate that the proposed method substantially outperforms existing temporal SOTA methods. On CVIS, the proposed method reduces mean error from 3.80 m to 1.57 m and improves R@1 m by 32.08 percentage points. On KITTI-CVL, direct transfer reduces the mean error from 3.57 m to 2.61 m, while target-domain fine-tuning further reduces it to 2.27 m and improves R@1 m to 35.69%. During real-vehicle field experiments, the proposed method achieves a mean error of 2.84 m, a median error of 2.92 m, and R@5 m of 96.86%, indicating effective cross-view localization on real roads where low-accuracy GPS measurements provide only local priors. These results show that the proposed temporal-context-enhanced cross-view localization framework provides an effective, generalizable, and practical solution for continuous visual localization.

Future work includes three directions. First, multimodal sensor cues, such as LiDAR, IMU, or HD map priors, can be integrated to improve robustness under heavy occlusion, severe illumination variation, and dynamic traffic. Second, motion constraints or trajectory-smoothing mechanisms can be incorporated into future temporal modeling to further improve long-horizon trajectory-level consistency. Third, more efficient temporal feature modeling and lightweight deployment strategies can reduce the additional cost introduced by continuous attention, better satisfying real-time and resource-constrained onboard localization requirements.